# Extreme Learning Machine for land cover classification

By

Mahesh Pal

# Extreme Learning Machine for land cover classification

Mahesh Pal
Department of Civil Engineering
N. I. T. Kurukshetra, Harayana, 136119, INDIA
mpce_pal@yahoo.co.uk, 01744 233356, Fax: 01744 238050

**Abstract:** This paper explores the potential of extreme learning machine based supervised classification algorithm for land cover classification. In comparison to a backpropagation neural network, which requires setting of several user-defined parameters and may produce local minima, extreme learning machine require setting of one parameter and produce a unique solution. ETM+ multispectral data set (England) was used to judge the suitability of extreme learning machine for remote sensing classifications. A back propagation neural network was used to compare its performance in term of classification accuracy and computational cost. Results suggest that the extreme learning machine perform equally well to back propagation neural network in term of classification accuracy with this data set. The computational cost using extreme learning machine is very small in comparison to back propagation neural network.

## 1. Introduction

Within the last two decades, neural network classifiers, particularly the feed-forward multi-layer perceptron using back-propagation algorithm, have been extensively tested for different applications by remote sensing community (Benediktsson et al., 1990; Hepner et al., 1990; Heermann and Khazenie, 1992; Civco, 1993; Schaale and Furrer, 1995; Tso and Mather, 2001). The popularity of neural network based classifiers is due to their ability to learn and generalize well with test data. In particular, neural networks make no prior assumptions about the statistics of input data. This property makes neural networks an attractive solution to many land cover classification of remotely sensed data whose underlying distribution is quite often unknown. The multilayer feed forward network is one of the most widely used neural network architectures. Among the various learning algorithms, the error backpropagation algorithm is one of the most important and widely used algorithms in remote sensing. A number of studies have reported that use of back propagation neural classifier have problems in setting various parameters during training (Kavzoglu, 2001; Wilkinson, 1997). The choice of network architecture (i.e. number of hidden layers and nodes in each layer, learning rate as well as momentum), weight initialisation and number of iterations required for training are some of the important parameters the affects the learning performance of these classifiers. The other shortcomings of the conventional backpropagation learning algorithm are slow convergence rate and it can get stuck to a local minimum.

Recently, Huang et al., (2006) proposed extreme learning machine based classification approach (also called as single hidden layer feed forward neural network) with randomly assigned input weights and bias. They suggested that this classification approach may not require adjusting the input weights like a backpropagation method and found it working well with different data set in comparison to back propagation neural network. Keeping this in view present study compares the performance of extreme learning machine with a back propagation neural network using multispectral data.

## 2.0 Extreme Learning Machine

Let the training data with $K$ number of samples be represented by $\{\mathbf{x}_i, \mathbf{y}_i\}$, where $\mathbf{x}_i \in \mathbf{R}^p$ and $\mathbf{y}_i \in \mathbf{R}^q$, a standard single hidden layer feed forward neural network (Huang and Babri, 1997) having $H$ hidden neurons and activation function $f(x)$ can be represented as:

$$\sum_{i=1}^{H} \alpha_i f_i(\mathbf{x}_j) = \sum_{i=1}^{H} \alpha_i f(\mathbf{w}_i \cdot \mathbf{x}_j + c_i) = \mathbf{e}_j \quad \text{where } j=1, \ldots, K, \tag{1}$$

Where $\mathbf{w}_i$ and $\alpha_i$ are the weight vectors connecting inputs and the $i$th hidden neurons and the $i$th hidden neurons and output neurons respectively, $c_i$ is the threshold of the $i$th hidden neuron and $\mathbf{e}_j$ is the output from single hidden layer feed forward neural network (SHLFN) for the data point $j$. The weight vector $\mathbf{w}_i$ is randomly generated and based on a continuous probability distribution (Huang et al., 2006).

Huang et al., (2006) suggested that a standard single layer feed forward neural network with $H$ hidden neurons and activation function $f(x)$ can approximate $K$ training data with zero error means such that:

$$\sum_{j=1}^{K} \|\mathbf{e}_j - \mathbf{o}_j\| = 0 \tag{2}$$

And the equation (1) can be expressed as

$$\sum_{i=1}^{H} \alpha_i f(\mathbf{w}_i \cdot \mathbf{x}_j + c_i) = \mathbf{o}_j, \quad j = 1, \ldots, K \tag{3}$$

for particular values of $\alpha_i, \mathbf{w}_i, \text{and } c_i$.

Further, Huang et al., (2006) proposed that equation (3) can be written in a compact form and represented by the following equation:

$$\mathbf{A}\alpha = \mathbf{Y} \tag{4}$$

Where $A$ is called the hidden layer output matrix of the neural network (Huang and Babri, 1997) and defined as:

$$A(\mathbf{w}_1, \ldots, \mathbf{w}_H, c_1, \ldots, c_H, \mathbf{x}_1, \ldots, \mathbf{x}_K) = \begin{bmatrix} f(\mathbf{w}_1 \cdot \mathbf{x}_1 + c_1) & \cdots & f(\mathbf{w}_H \cdot \mathbf{x}_1 + c_H) \\ \vdots & & \vdots \\ f(\mathbf{w}_1 \cdot \mathbf{x}_K + c_1) & \cdots & f(\mathbf{w}_H \cdot \mathbf{x}_K + c_H) \end{bmatrix}_{K \times H} \tag{5}$$

$$\alpha = \begin{bmatrix} \alpha_1^T \\ \vdots \\ \alpha_H^T \end{bmatrix}_{H \times m} \quad \text{and} \quad \mathbf{Y} = \begin{bmatrix} \mathbf{y}_1^T \\ \vdots \\ \mathbf{y}_K^T \end{bmatrix}_{K \times m} \tag{6}$$

The $i$th column of $A$ is the $i$th hidden neuron's output vector with respect to inputs $x_1, x_2, \ldots, x_K$.

The SHLFN can be solved by using a gradient based solution and one need to find the suitable values of $\mathbf{w}_i', c_i' \text{ and } \alpha' \, (i=1, \ldots, H)$ such that

$$\left\| A(\mathbf{w}_1', \ldots, \mathbf{w}_H', c_1', \ldots, c_H') \alpha' - \mathbf{Y} \right\| = \min_{\mathbf{w}_i, c_i, \alpha} \left\| A(\mathbf{w}_1, \ldots, \mathbf{w}_H, c_1, \ldots, c_H) \alpha - \mathbf{Y} \right\| \tag{7}$$

Equation (7) can be written in form of the following cost function

$$C = \sum_{j=1}^{K}\left(\sum_{i=1}^{H}\alpha_i\ f(\mathbf{w}_i \cdot \mathbf{x}_j + c_i) - \mathbf{o}_j\right)^2 \quad (8)$$

which can be minimised to find suitable values of $\mathbf{w}_i'$, $c_i'$ and $\alpha'$ $(i=1,......,H)$.

In case the hidden layer output matrix of neural network (i.e. $A$) is unknown, a gradient based learning algorithm minimise the $\|A\alpha - Y\|$ by adjusting a vector $W$ (i.e. a set of *weights* $(\mathbf{w}_i, \alpha_i)$ *and biases* $(c_i)$) iteratively by using the following relationship:

$$\mathbf{W}_p = \mathbf{W}_{p-1} - \eta\frac{\partial C(\mathbf{W})}{\partial \mathbf{W}} \quad (9)$$

where $\eta$ is the learning rate and backpropagation learning algorithms is one of the most popular algorithm used to compute the gradients in a feed forward neural network.

Recently, the study carried out by Huang et al., (2003) proved that single layer feed forward neural network with randomly assigned input weights and hidden layer biases and with almost any nonzero activation function can universally approximate any continuous functions on any input data sets. Huang et al., (2006) suggested an alternate way to train a SHLFN by finding a least square solution $\alpha'$ of the linear system represented by equation 4:

$$\|\mathbf{A}(\mathbf{w}_1,......,\mathbf{w}_H, c_1,........,c_H)\alpha' - \mathbf{Y}\| = \min_{\alpha}\|\mathbf{A}(\mathbf{w}_1,......,\mathbf{w}_H, c_1,........,c_H)\alpha - \mathbf{Y}\| \quad (10)$$

If the number $H = K$, matrix $A$ is square and invertible but in most of the cases number of hidden nodes are less than the number of training samples, which makes matrix $A$ to be a non square matrix and there may not exist $\mathbf{w}_i, c_i, \alpha_i$ $(i=1,.......H)$ such that $\mathbf{A}\alpha = \mathbf{Y}$. To overcome this problem, Huang et al., (2006) proposed in using smallest norm least squares solution of $\mathbf{A}\alpha = \mathbf{Y}$, thus, the solution of equation 4 becomes:

$$\alpha' = \mathbf{A}^{@}\mathbf{Y} \quad (11)$$

Where $\mathbf{A}^{@}$ is called Moore-Penrose generalized inverse of matrix $A$ (Serre, 2002). This solution has the following important properties (Huang et al., 2006):

1. The smallest training error can be reached by this solution.
2. Smallest norm of weights and best generalization performance.
3. The minimum norm least-square solution is a unique solution, thus involving no local minima like one in backpropagation learning algorithm.

Thus, the algorithm proposed by Huang et al., (2006) and called as extreme learning machine can be summarized as:

With the training data set $\{\mathbf{x}_i, \mathbf{y}_i\}$, $\mathbf{x}_i \in \mathbf{R}^n$, $\mathbf{y}_i = \in \mathbf{R}^m$ having $K$ number of samples, a standard SHLFN algorithm with $H$ hidden neurons and activation function $g(x)$ will work as:

1. Assign random input weights $\mathbf{w}_i$ and bias $c_i$, $i=1,.......H$.
2. Calculate the hidden layer output matrix $A$.
3. Calculate the output weights $\alpha$ by using the following equation

$$\alpha = \mathbf{A}^{@}\mathbf{Y}$$

Where $A$, $\alpha$ and $Y$ are as defined in equations 5 and 6.

## 3. Data Sets and Methodology

The study areas used in this study is located near the town of Littleport in eastern England and the image was acquired on 19 June 2000. A sub-image consisting of 307-pixel (columns) by 330-pixel (rows) covering the area of interest was used for subsequent analysis and classification problem involved in identification of seven land cover types (i.e. wheat, potato, sugar beet, onion, peas, lettuce and beans). A total of 4737 pixels were selected for all seven classes using stratified random sampling. The pixels collected were divided into two subsets, one of which was used for training and the second for testing the classifiers, so as to remove any bias resulting from the use of the same set of pixels for both training and testing. Also, because the same test and training data sets are used for each classifier, any difference resulting from sampling variations was avoided. A total of 2700 training and 2037 test pixels were used.

## 4. Results

The purpose of the present study is to evaluate the performance of extreme learning machine for land cover classification and comparing its performance with a back propagation neural network classifier. Unlike back propagation neural network, the design of extreme learning machine requires setting of one user-defined parameter i.e. number of hidden nodes in hidden layer. A number of experiments were carried out by using the training and test data set of 2700 pixels and varying the hidden nodes from 25 to 450. Results suggests that extreme learning machine achieves highest classification accuracy with a total of 300 hidden nodes.

**Table1. Classification accuracy, user-defined parameters as well computational cost with both classifiers.**

| Classifier used | User defined parameters | Accuracy (%) | Computational cost (seconds) |
|---|---|---|---|
| Extreme learning machine | Number of hidden nodes = 300 | 89.0 | 1.25 |
| Back propagation neural network | Learning rate =0.25, Momentum = 0.2, nodes in hidden layers =26, number of iterations = 2200, number of hidden layers =1 | 87.75 | 336.20 |

Table 1 provides the results obtained by using extreme learning machine as well as back propagation neural network using ETM+ (England) data set. Results suggest that extreme learning machine perform well in comparison to the back propagation neural network. With this dataset, extreme learning machine provide a classification accuracy of 89% in comparison to 87.87% by a back propagation neural network. Computational cost (i.e. training and test time) of a classifier often represents a significant proportion of cost in remote sensing classifications. For all experiments in this study, a personal computer with a Pentium IV processor and 512 MB of RAM was used. Table 1 also provide the computational cost using ETM+ data set with extreme learning machine and back propagation neural network. The results (table 1) suggest the usefulness of extreme learning machine in comparison to back propagation neural network in term of computational cost also.

## 5.0 Conclusions

The main aim of this study was to assess the usefulness of extreme learning machine based classification approach for land cover classification using multispectral data. The performance of extreme learning machine was compared with a back propagation neural network. The results presented above suggest that extreme learning machine works equally well to back propagation neural network in term of classification accuracy and involves in using a smaller computational cost. Another conclusion about the use extreme learning machine classification approach is that unlike a back propagation neural network classifier its performance is affected by one user-defined parameter only which can easily be identified for a particular data set.

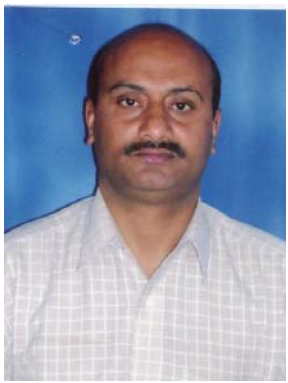

Mahesh Pal is currently working as assistant professor in the department of civil engineering, NIT Kurukshetra, Haryana. He has 35 Paper in international/national journals and conferences. His current interests include kernel based approaches for land cover classification and application of GIS in construction management.